\documentclass[10pt,twocolumn,letterpaper]{article}

\usepackage{iccv}
\usepackage{times}
\usepackage{epsfig}
\usepackage{graphicx}
\usepackage{amsmath}
\usepackage{amssymb}

\usepackage{algorithm}
\usepackage{algpseudocode}
\usepackage{bm}
\usepackage{subcaption}

\usepackage[breaklinks=true,bookmarks=false]{hyperref}

\iccvfinalcopy 

\def\iccvPaperID{****} 

\ificcvfinal\fi

\begin{document}

\title{GST: Group-Sparse Training for Accelerating Deep Reinforcement Learning}

\author{Juhyoung Lee \qquad Sangyeob Kim \qquad Sangjin Kim \qquad Wooyoung Jo \qquad Hoi-Jun Yoo \\
School of Electrical Engineering\\
KAIST, Daejeon, Republic of Korea\\
{\tt\small \{juhyoung, sangyeob.kim, sangjinkim, jobin2725, hjyoo\}@kaist.ac.kr}
}

\maketitle
\ificcvfinal\thispagestyle{empty}\fi

\begin{abstract}
Deep reinforcement learning (DRL) has shown remarkable success in sequential decision-making problems but suffers from a long training time to obtain such good performance. Many parallel and distributed DRL training approaches have been proposed to solve this problem, but it is difficult to utilize them on resource-limited devices. In order to accelerate DRL in real-world edge devices, memory bandwidth bottlenecks due to large weight transactions have to be resolved. However, previous iterative pruning not only shows a low compression ratio at the beginning of training but also makes DRL training unstable. To overcome these shortcomings, we propose a novel weight compression method for DRL training acceleration, named group-sparse training (GST). GST selectively utilizes block-circulant compression to maintain a high weight compression ratio during all iterations of DRL training and dynamically adapt target sparsity through reward-aware pruning for stable training. Thanks to the features, GST achieves a 25 \%p $\sim$ 41.5 \%p higher average compression ratio than the iterative pruning method without reward drop in Mujoco Halfcheetah-v2 and Mujoco humanoid-v2 environment with TD3 training.
\end{abstract}

\section{Introduction}

Deep reinforcement learning (DRL) algorithms have made remarkable success in sequential decision-making problems like gaming agents, autonomous robots, and human-computer interaction. Adopting deep neural network (DNN) to approximate the policy or value estimator of RL, DRL algorithms have overcome the performance limitation of classical RL algorithms and have achieved the human-level or even better performance in various large-scale models or environments. According to recent researches of Deepmind \cite{humanlevel, gamego, gamestar2}, the trained DRL agents have overwhelmed skilled human players from simple video games like Atari to complex games like Go and StarCraft II.

However, it is very difficult to obtain such high-performance DRL agents because a huge amount of data is required to train the neural networks utilized for DRL.  Unlike supervised learning in which labels exist, DNN in DRL inevitably suffers from unstable training because it should be trained with experience data obtained through interaction between the untrained DNN and the environment. The unstable training causes frequent communication to the environment and a lot of model parameter updates, which leads to a large training time. \cite{gamestar2} trained a Starcraft II DRL agent, using 32 tensor processing units (TPUs) over 44 days.

There are several studies for reducing the overall training time required for DRL \cite{A3C, GA3C, impala, seedrl}. The biggest problem of standard DRL implementations, including A2C and PPO \cite{PPO}, is a significant under-utilization of computing resources caused by serial execution of experience sampling and computing neural networks. Most of the previous work tried to handle the limitation through a parallel and distributed DRL framework. A3C \cite{A3C} utilized independent actors with independent policies that perform environment simulation, action generation, and gradient calculation in parallel. GA3C \cite{GA3C} outperformed CPU-only A3C by adopting GPU. IMPALA \cite{impala} improved the GA3C architecture through efficient GPU batching and Seed RL \cite{seedrl} achieved higher throughput compared to \cite{impala} by reducing communication time. By accelerating experience sampling in parallel through distributed agents, the above frameworks have achieved state-of-the-art performance in DRL, such as shortening the training time of Atari games to several hours.

However, it is hard to utilize parallel and distributed DRL algorithms in resource-constrained devices like mobile or edge devices due to their large amount of computational workload. Those algorithms require a vast amount of computing resources including hundreds of CPU cores and DNN accelerators for experience sampling from parallel environments and fast DNN training. Seed-RL \cite{seedrl}, the state-of-the-art DRL framework, utilized dozens to hundreds of CPUs and 8 TPU v3 cores to accelerate the entire training of DRL. As more and more recent researches highlight a real-world DRL \cite{realworldrl, challengesrealworldrl} and adaptation to the sudden environment change of edge devices through DRL \cite{learntoadapt}, efficient methods for DRL training acceleration on resource-constrained devices are essential.

\begin{figure} [t]
\begin{center}
   \includegraphics[width=0.85\linewidth]{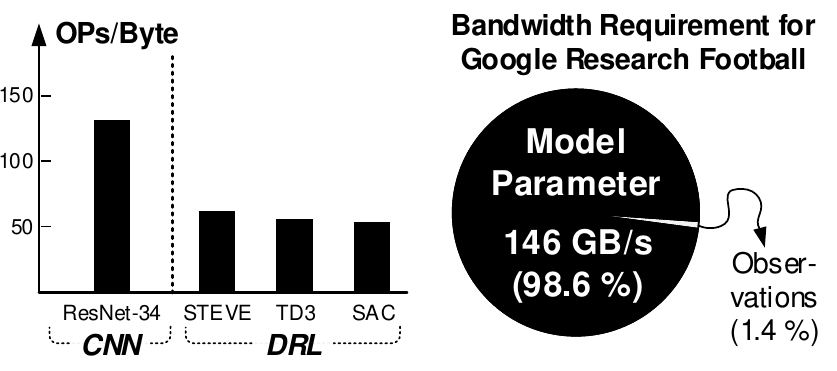}
\end{center}
\vspace*{-3mm}
   \caption{Large memory bandwidth requirement of DRL.}
\label{fig:requirementofDRL}
\end{figure}

The biggest problem of DRL on mobile devices is memory bandwidth bottlenecks. Unlike previous DRL frameworks, mobile DRL frameworks utilize only a few parallel environments or do not use simulation environment. Therefore, the training speed in mobile DRL platforms is limited by memory bandwidth bottleneck due to DNN training rather than by the parallelism of the simulation environment. Figure~\ref{fig:requirementofDRL} shows two main reasons for large memory access of DRL training; (1) DNNs utilized for DRL are mainly composed of fully-connected (FC) layer or recurrent layer which require a lot of memory access for high throughput due to small operational intensity. Compared to ResNet-34, which shows more than 100 Ops/byte operational intensity, the operational intensity of several famous DRL algorithms (STEVE \cite{steve}, TD3\cite{TD3}, SAC\cite{haarnoja2018soft}) for Mujoco environment is limited to 50 $\sim$ 60 Ops/byte. (2) The complex and sequential execution of multiple DNNs in DRL requires frequent access to model parameters and experiences, resulting in large memory access. Especially, the bandwidth required by the model parameter is much larger than the bandwidth required by the observation (98.6 \% in Google Research Football \cite{grf}).

The traditional method to solve the memory bottleneck caused by a large number of model parameters is model compression. In particular, model pruning shows state-of-the-art performance among model compression methods by removing unnecessary weight connections. However, it is difficult to utilize conventional pruning methods for the acceleration of DRL training due to two reasons. (1) Pruning should be performed iteratively during training to prevent severe performance drop, the sparsity of the model parameter starts at 0 and increases little by little. Therefore, it shows a low weight compression ratio at the beginning of training, which limits the average compression ratio of training. (2) Fixed sparsity scheduling method in previous iterative pruning \cite{gradualpruning} makes the DRL training unstable.

In this paper, we propose group-sparse training (GST), a training method that can overcome the limited compression ratio of previous pruning methods at early iteration. Compared with the previous compression methods which utilized only pruning for entire training iterations, GST selectively utilizes block-circulant grouping along with pruning. Specifically, in the early iteration of training, both block circulant and pruning are applied simultaneously to increase the compression ratio, and if sparsity is sufficiently high in the latter iteration of training, the block-circulant grouping is selectively released to prevent a reward degradation. Moreover, we propose reward-aware pruning (RWP), which dynamically schedules target sparsity according to reward history to achieve not only stable training but also a high compression ratio.

\begin{figure*} [!htb]
\begin{center}
   \includegraphics[width=0.85\linewidth]{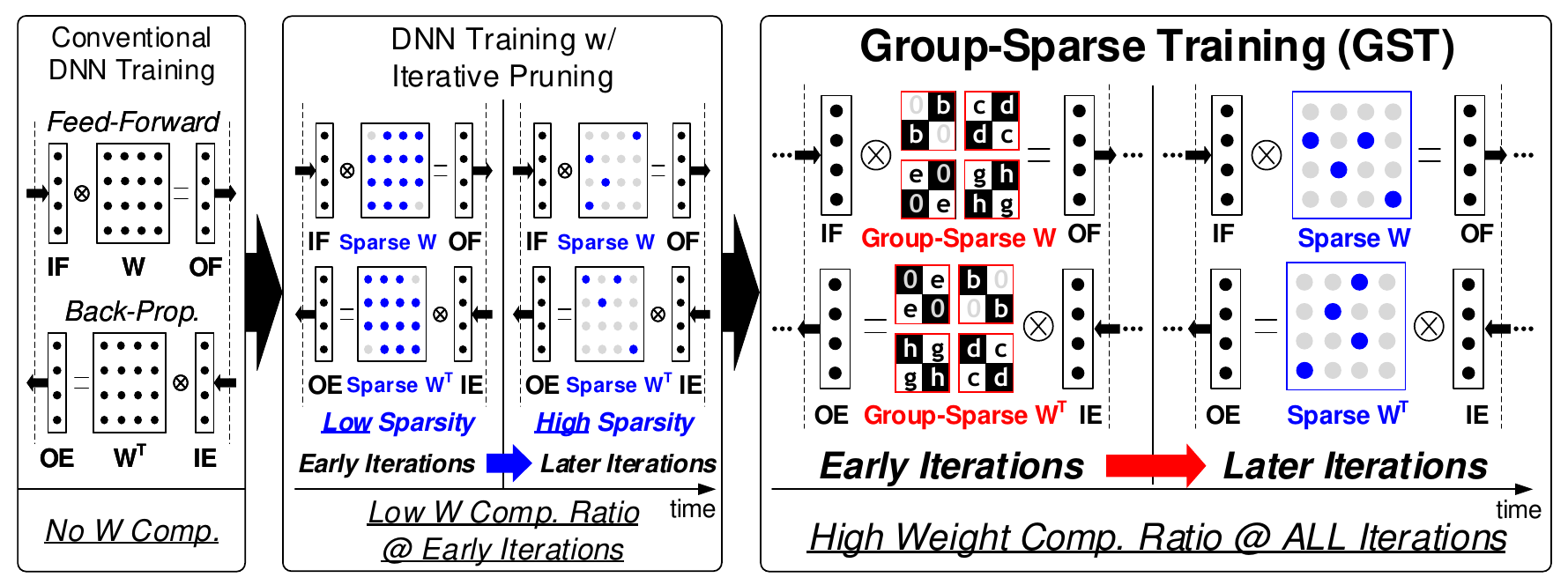}
\end{center}
\vspace*{-3mm}
   \caption{Overall flow of the proposed group-sparse training (GST).}
\label{fig:gstoverallflow}
\end{figure*}

\section{Related Works}

It is widely accepted that a more efficient and faster DNN model can be generated by removing the redundant weight from the model. In this section, we introduce previous works on model pruning, structured matrix, and pruning for DRL or fast training.

{\bf Model Pruning.} The key concept of model pruning is eliminating redundant connections or neurons of the target neural network. Most of the pruning algorithms achieve a high compression ratio without loss of accuracy by repetitive removing and finetuning a small weight in a pre-trained network \cite{songhanprune}. Gradual pruning \cite{gradualpruning} shows that iteratively removing the weight connection during network training can achieve a higher compression ratio. Unlike the above methods in which unstructured sparsity is induced, recent researches tried to create structured sparsity of filter-level or channel-level \cite{filterpruning}. Most pruning studies have focused on obtaining high sparsity for efficient DNN inference, and cannot accelerate the DNN training process. Indeed, their training processes are even slowed down due to the utilization of a pre-trained model or knowledge distillation.

{\bf Structured Matrix.} The structured matrix methods compress DNNs by expanding shared weights into a predetermined matrix format. \cite{structured_transform, circulant_early, acdc} first compressed fully-connected (FC) layers using a structured matrix format. CirCNN \cite{circnn} achieved not only a higher compression ratio by using a circulant-matrix format in a block unit but also significantly reduced the amount of computation by utilizing FFT. CircConv \cite{circonv} showed that the block-circulant method can be utilized for convolutional layers. Compared to the model pruning method, the structured matrix has a negligible cost for encoding but suffers from a lower compression ratio at the same accuracy.

{\bf Pruning for DRL or Fast Training.} There are only a few studies utilized model compression to accelerate DRL or training. Prunetrain \cite{prunetrain} first tried to accelerate training by using repeated regularization based pruning, but it suffered from a low compression ratio due to the low initial and final sparsity of the regularization based pruning method. PoPS \cite{pops} applied pruning to DRL for the first time. \cite{pops} was able to achieve a high compression ratio by repeatedly training a small student policy network using knowledge distillation, but it cannot accelerate the training process because of a pre-trained teacher network for pruning. \cite{acceldrl} accelerated the training of DRL by using the knowledge distillation method, but there was also a limitation in DRL training speed-up because of a large teacher network.

\section{Methodology}

The objective of the proposed Group-Sparse Training (GST) is to solve the memory bandwidth bottleneck due to large parameters by compressing the DNN models during all DRL training iterations. The proposed GST is designed to be included in the training loop of any DRL algorithms like deep Q learning (DQN) \cite{humanlevel}, TD3 \cite{TD3}, PPO \cite{PPO}, etc. Moreover, since GST only affects inference and backpropagation of DNN, DRL algorithms for better sample efficiency \cite{per} can be used together with the GST.

\begin{algorithm} [!htb]
\caption{Algorithm Description of GST}
\label{alg:gst_overall}
\textbf{Input:} environment $\bm{Env}$, pruning threshold step $\bm{P_{step}}$, pruning start step $\bm{P_{start}}$, pruning freqeuncy $\bm{P_{fre}}$, sparsity upper bound $\bm{S_{ub}}$, block size $\bm{B}$, Phase shift sparsity $\bm{S_{shift}}$, model parameter $\bm{W}$

\textbf{Output:} The compressed and trained model parameter $\bm{W^*}$ 

\textbf{Initialize:} initial model parameters $\bm{W}$, current pruning threshold $\bm{P_{th}}=0$, previous highest reward $\bm{R_{prev}}=0$, current reward $\bm{R_{new}}=0$, timestep $\bm{T}=0$, current sparsity $\bm{S_{now}}=0$ 
\begin{algorithmic}[1]
\If{($\bm{B}>1$) and ($\bm{S_{shift}}\neq0$)}
\State{Reinitialize $\bm{W}$ as block-circulant form (size $\bm{B}$)}
\EndIf
\For{$\bm{T}=0$; $\bm{T}=\bm{T_{max}}$; $\bm{T++}$}
\State{$\bm{next\_state}$, $\bm{R_{new}}$  = Exp\_Gen($\bm{Env, W, state}$)}
\State{Calculate gradient based on generated experience}
\State{Measure current model parameters’ sparsity $\bm{S_{now}}$}
\If{$\bm{S_{shift}}>\bm{S_{now}}$}
\State{Gradient gen. as block-circulant form (size $\bm{B}$)}
\EndIf
\State{Update the model parameter $\bm{W}$ based on gradient}
\If{($mod(\bm{T},\bm{P_{fre}})=0$) and ($\bm{P_{start}}<\bm{T}$)}
\If{$\bm{S_{now}}<\bm{S_{ub}}$}
\State{Apply Reward-Aware Pruning}
\EndIf
\EndIf
\If{$\bm{R_{new}}> \bm{R_{prev}}$}
\State{$\bm{R_{prev}}= \bm{R_{new}}$}
\EndIf
\EndFor
\end{algorithmic}
\end{algorithm}

\begin{algorithm} [!htb]
\caption{Algorithm Description of RWP}
\label{alg:gst_rwp}
\textbf{Input:} pruning threshold step $\bm{P_{step}}$, previous highest reward $\bm{R_{prev}}$, current reward $\bm{R_{new}}$, model parameters $\bm{W}$, current pruning threshold $\bm{P_{th}}$

\textbf{Output:} The pruned model parameter $\bm{W_{pruned}}$, Updated pruning threshold $\bm{P_{th\_new}}$ 
\begin{algorithmic}[1]
\If{$\bm{R_{new}}>\bm{R_{prev}}$}
\State{$\bm{P_{th\_new}}= \bm{P_{th}}+\bm{P_{step}}$}
\State{Zeroize bottom $\bm{P_{th\_new}}$\% of each layer’s weight}
\EndIf
\end{algorithmic}
\end{algorithm}

\subsection{The Architecture of the Group-Sparse Training}
The basic concept of GST is the selective utilization of block-circulant weight compression to compensate for an insufficient compression ratio of iterative pruning in the early iterations of training.

Figure~\ref{fig:gstoverallflow} shows the overall flow of the proposed GST and a comparison between the GST and the previous training method. Conventional DNN training is consists of a feed-forward process with uncompressed weight ($W$), and a back-propagation process with uncompressed transposed weight ($W^T$).  In the iterative pruning algorithm, sparsity is created by gradually removing connections with small absolute values from $W$ and $W^T$ as training progresses. However, the iterative pruning algorithm can remove only a small amount of weights at the beginning of training because it cannot determine the saliency of the specific connection before sufficient training is performed. Low sparsity at the early iterations causes two problems; (1) lower the average compression ratio over the entire iteration; (2) the large compression ratio difference between the early and the later iterations makes hard to optimize memory bandwidth design for weight transaction.

\begin{figure}
\begin{center}
   \includegraphics[width=0.88\linewidth]{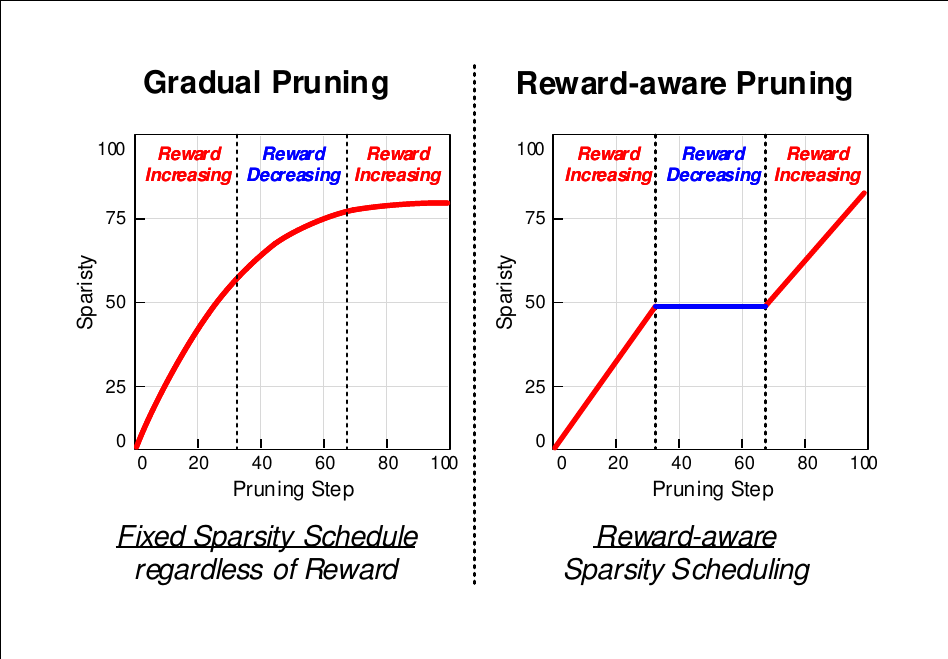}
\end{center}
\vspace*{-3mm}
\caption{Comparison between gradual pruning \cite{gradualpruning} and the proposed reward-aware pruning (RWP).}
\label{fig:GPRWPcomp}
\end{figure}

\begin{figure}
\begin{center}
   \includegraphics[width=0.95\linewidth]{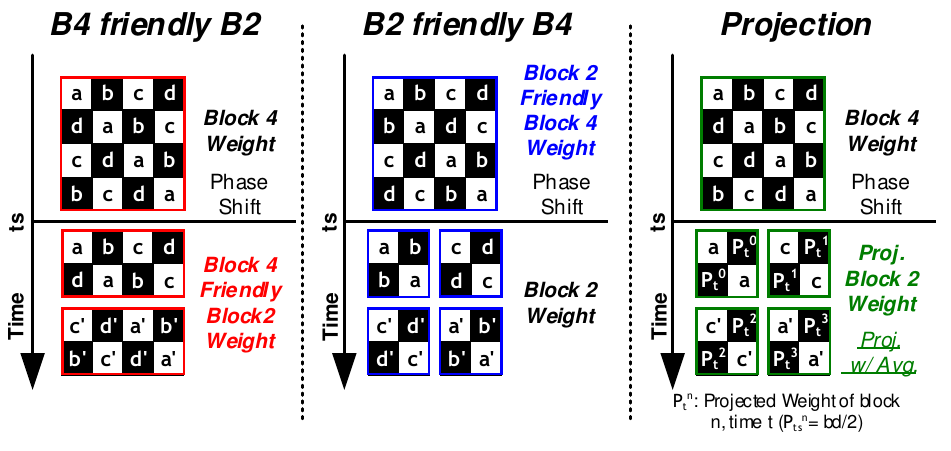}
\end{center}
\vspace*{-3mm}
   \caption{Block size conversion methods for supporting multiple block sizes during training.}
\label{fig:blockchange}
\end{figure}

To mitigate the limitation of iterative pruning, the proposed GST selectively utilizes a block-circulant weight compression. Compared with iterative pruning, the block-circulant weight compression method not only shows the negligible cost for index encoding but also can be applied to all training iterations. However, its compression ratio should be limited to prevent severe training performance drop because it cannot determine redundant weight like iterative pruning. During the early iterations, GST groups the weights in a block-circulant form while pruning the grouped weights simultaneously. We call the weight at this stage a group-sparse weight. Only after the sparsity of weights gets sufficiently high, GST selectively unpacks the grouped weights and iteratively prune the weights until completion. By transitioning between the group-sparse and sparse weight compression according to the progress of training, GST can achieve a higher weight compression ratio without sacrificing the training performance.

The detailed description of the proposed GST is in Algorithm~\ref{alg:gst_overall}. Note that the comparison result between the pre-define phase shift sparsity and the current sparsity determines whether to apply the block-circulant. The phase shift algorithm is motivated by the fact that the reward limitation due to block-circulant begins after a certain amount of sparsity is obtained. By appropriately disabling the block-circulant, it is possible to achieve a high compression ratio without training performance loss. Results of phase shift algorithms are described in the experiment section.

\subsection{Reward-aware Pruning}
In addition to the GST, we developed reward-aware pruning (RWP), an iterative pruning method that can minimize the reduction of reward in the DRL domain. The key concept of RWP is to enable stable training by dynamically adjusting the target sparsity of pruning according to the history of reward values.

In the previous iterative pruning algorithm \cite{gradualpruning} targeting classification networks, the gradual pruning function (\ref{eq:gradualpruning}) was utilized to determine the target sparsity according to training iterations.
\begin{equation} \label{eq:gradualpruning}
    s_t=s_f+(s_i-s_f)(1-{\frac{t-t_0}{n{\Delta}}})^3
\end{equation}
\noindent
where $s_i$ is initial sparsity, $s_f$ is the final sparsity, $t_0$ is pruning start point, ${\Delta}$ is pruning frequency, $n$ is pruning steps.

However, as pointed out in \cite{pops}, applying (\ref{eq:gradualpruning}) to DRL training can make the training procedure highly unstable. This is because equation (\ref{eq:gradualpruning}) manages the pruning ratio in a fixed manner regardless of the status of training. Compared with classification DNN training, DRL training shows an unstable training curve and has various convergence speeds depending on the task. Therefore, pruning with a fixed sparsity schedule cannot sufficiently consider the saliency of the weight parameter in DRL, which leads to a training performance drop. Previous pruning papers in DRL domains \cite{pops, acceldrl} suggested network compression based on knowledge distillation to overcome this problem, but they are hard to accelerate DRL training procedure due to utilization of pre-trained or large teacher network.

Figure~\ref{fig:GPRWPcomp} shows the proposed RWP and previous gradual pruning \cite{gradualpruning}. The proposed RWP dynamically changes target sparsity according to the reward. Specifically, RWP increases target sparsity only when the obtained reward is higher than the previous highest reward. By using RWP, it is possible to dynamically find a pruning pattern suitable for various convergence patterns of DRL. The RWP is summarized in Algorithm~\ref{alg:gst_rwp}.

\subsection{Block Size Conversion Methods}

Instead of controlling only the application of block-circulant during training, the GST can control block size $B$ during training to achieve a higher compression ratio. Compared with utilizing only small $B$ for the entire training, the compression ratio can be further increased without training performance loss by utilizing a large $B$ for early iterations of training and a small $B$ for later iterations of training. However, it is impossible to change the block size $B$ directly during training because there is no compatibility between different sized block-circulant matrixes. Therefore, methods for converting a weight trained with a large $B$ to a weight that is trainable with a small $B$ is needed.

Figure~\ref{fig:blockchange} shows three ways to convert $B=4$ weight to $B=2$ weight; (1) projection, (2) block4 friendly block2, (3) block2 friendly block4. The most simple conversion method is projecting the weight to the target $B$ circulant tensor \cite{circonv} when a block size transition occurs. However, the projection method causes a huge training performance drop due to a sudden parameter value change after the block size transition and makes subsequent training unstable. The block4 friendly block2 method composes the block2 structure by dividing the block4 structure in half instead of changing the parameter value. The block2 friendly block4 method composes the block4 structure as a collection of 4 block2 structures. By utilizing the friendly grouped matrix, the block size transition can be performed without changing the parameter value to prevent a training performance drop that occurs in the transition. The results of the three methods are described in the experiment section.

\subsection{Estimation of Compression Ratio}
\label{section:crcompute}

This paper proposes a weight compression algorithm for a system that can support block-circulant and unstructured sparsity through an ASIC implementation. Ideally, the compression ratio of the DNN with sparsity $S$ and block size $B$ is expressed by the following equation (\ref{eq:compratio}).
\begin{equation} \label{eq:compratio}
    CR=(\frac{B+S-1}{B}){\times}\frac{P_{comp}}{P_{total}}
\end{equation}
\noindent
where $P_{total}$ is the total number of parameters and $P_{comp}$ is the number of parameters to which GST is applied.

However, even if ASIC implementation is considered, the overhead for encoding information of unstructured sparsity should be considered. In this paper, bitmap sparsity encoding is adopted to estimate the overhead. The bitmap method encodes sparsity with a bitmap whose value is 1-bit 0 at the zero weight position and 1-bit 1 at the non-zero weight position. The other encoding methods like CSR or Viterbi \cite{viterbi} showed a higher compression ratio than the bitmap for high sparsity, but their encoding overhead is greater than the bitmap for low sparsity. The compression ratio of the DNN with 16-bit weight and bitmap encoding is expressed by the following equation (\ref{eq:advanedcompratio}).
\begin{equation} \label{eq:advanedcompratio}
    CR=(\frac{B+S-1}{B}-\frac{1}{16}){\times}\frac{P_{comp}}{P_{total}}
\end{equation}

\begin{figure*} [!htb]
    \begin{center}
    \begin{subfigure}{0.4\textwidth}
        \includegraphics[width=\linewidth]{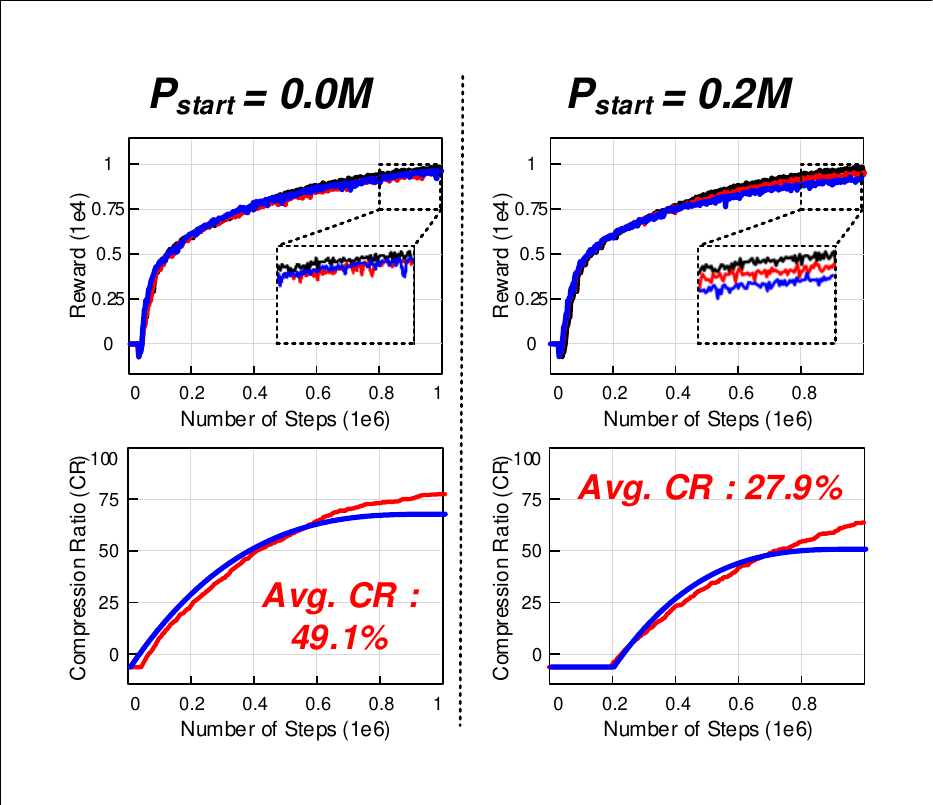}
        \caption{Result on mujoco halfcheetah-v2 with TD3 \cite{TD3}} \label{fig:5a}
    \end{subfigure}%
    \hspace*{1cm}   
    \begin{subfigure}{0.4\textwidth}
        \includegraphics[width=\linewidth]{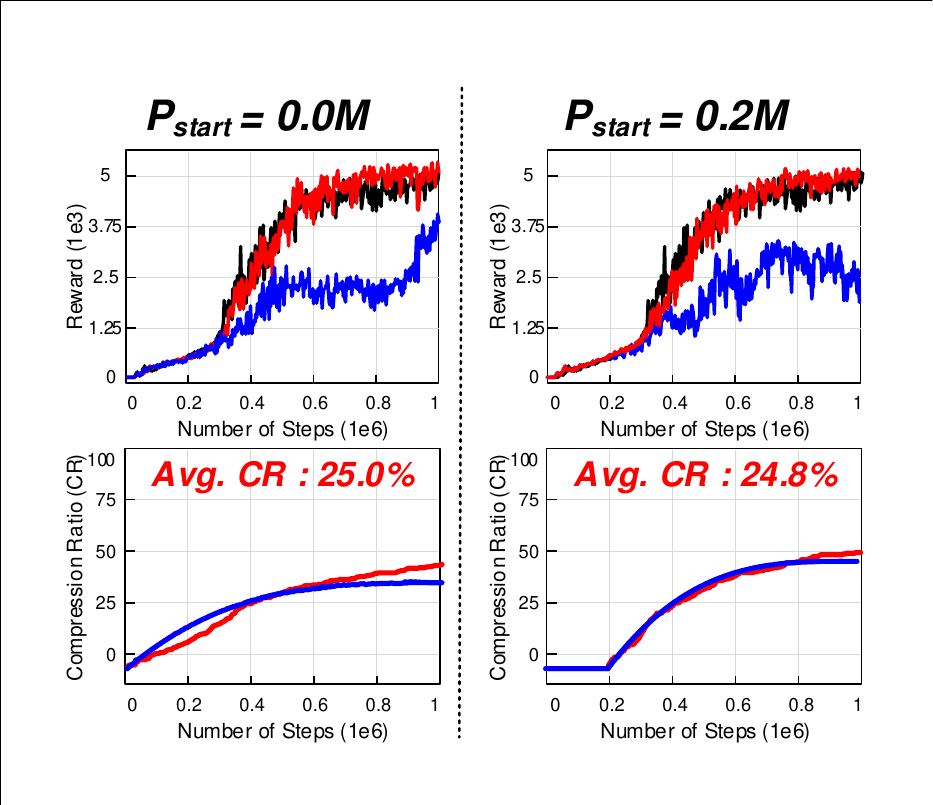}
        \caption{Result on mujoco humanoid-v2 with TD3 \cite{TD3}} \label{fig:5b}
    \end{subfigure}%
    \end{center}
    \vspace*{-3mm}
   \caption{Reward and compression ratio measurement results of gradual pruning \cite{gradualpruning} and reward-aware pruning. Black lines are baseline result (No pruning), blue lines are gradual pruning result, and red lines are reward-aware pruning results.}
\label{fig:resultGPvsRWP}
\end{figure*}

\begin{figure*} [!htb]
\begin{center}
   \includegraphics[width=0.8\linewidth]{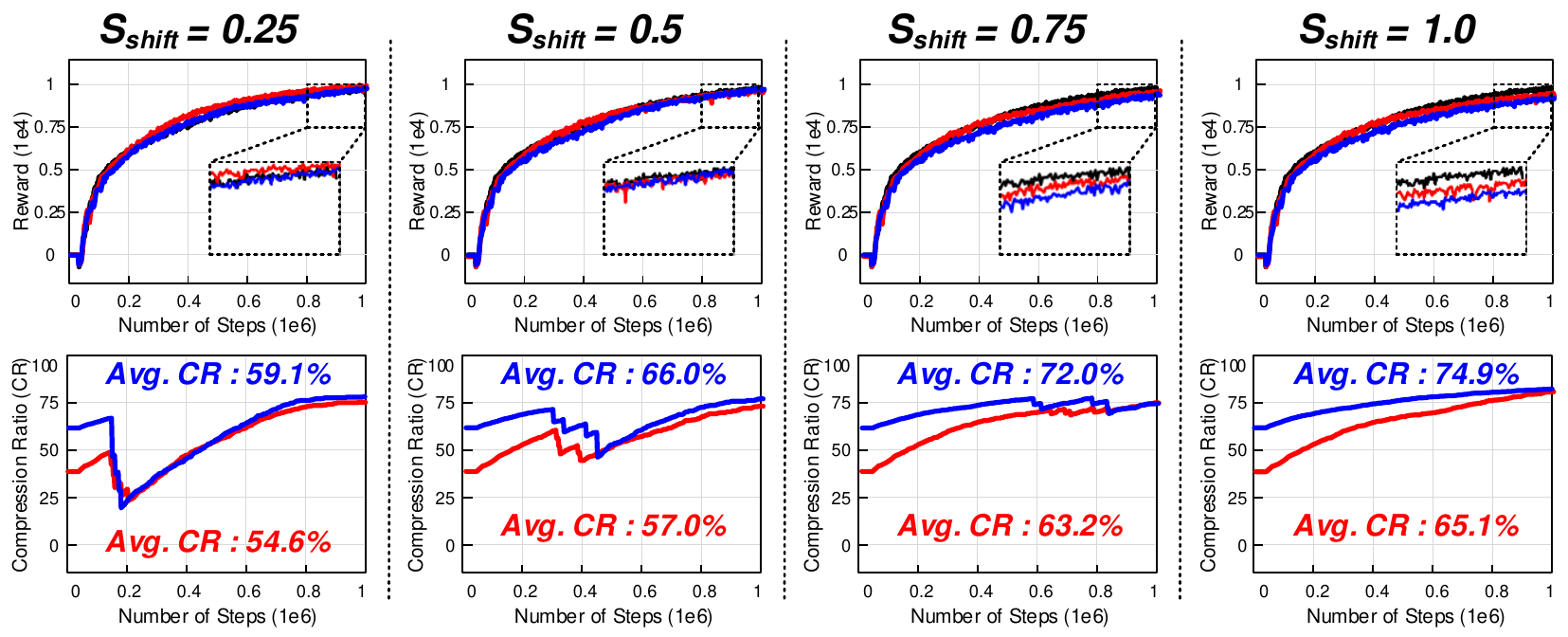}
\end{center}
\vspace*{-3mm}
   \caption{Reward and compression ratio measurement results according to different block size and phase shift sparsity. Measurement environment is mujoco halfcheetah-v2 with TD3 \cite{TD3}. Black lines are baseline result, blue lines are block size 4 results, and red lines are block size 2 results.}
\label{fig:resultHCGST}
\end{figure*}

\begin{figure} [!htb]
\begin{center}
   \includegraphics[width=0.8\linewidth]{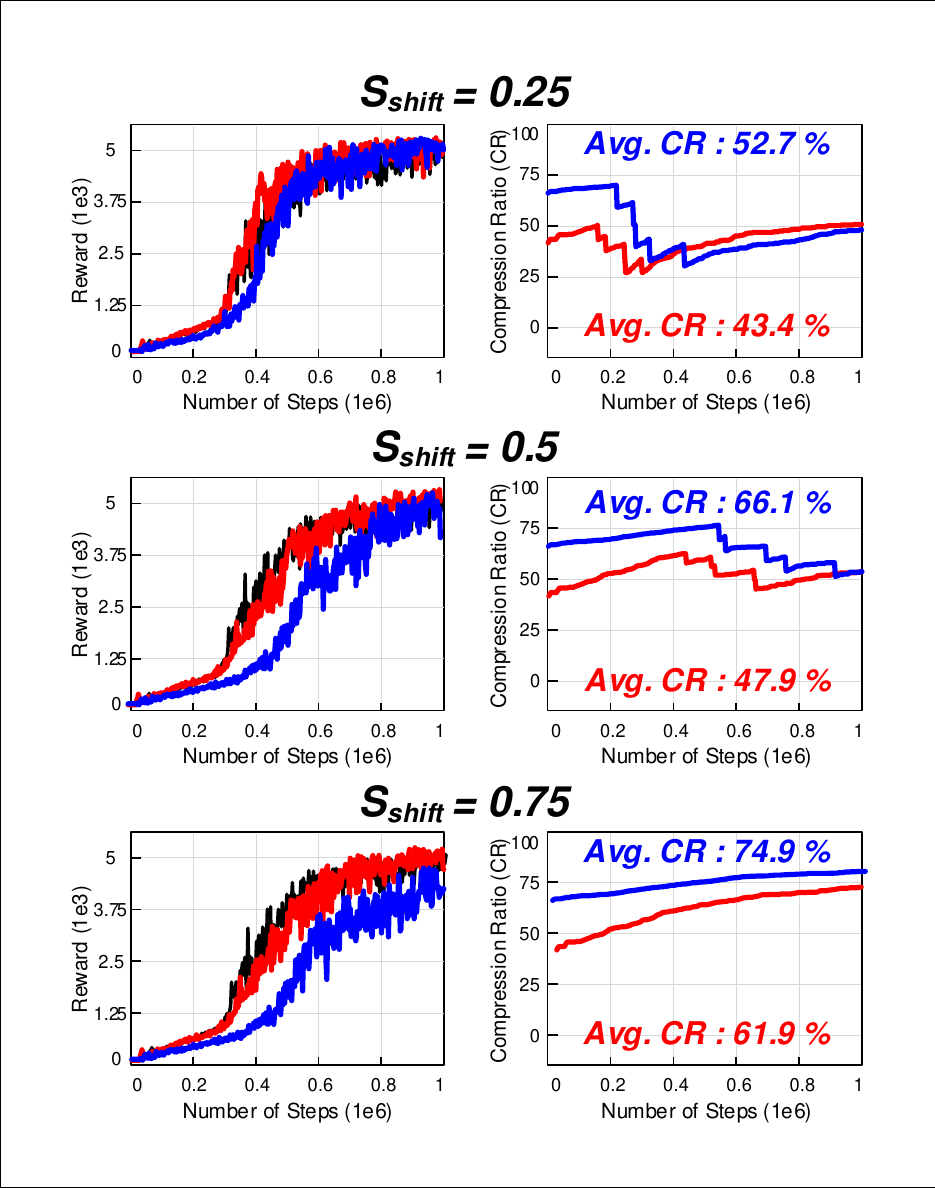}
\end{center}
\vspace*{-3mm}
   \caption{Reward and compression ratio measurement results according to different block size and phase shift sparsity. Measurement environment is mujoco humanoid-v2 with TD3 \cite{TD3}. A line color configuration is the same as the configuration of Figure~\ref{fig:resultHCGST}.}
\label{fig:resultHUGST}
\end{figure}
\section{Experiments}
In this section, we compare the reward of the proposed training algorithm and the conventional training algorithm while measuring the obtained compression ratio (CR). In order to show that the proposed GST can compress weight parameters while maintaining training performance without additional training iterations, the entire training curves were measured and reported. The CRs were measured with the method described in equation (\ref{eq:advanedcompratio}).

The proposed GST can be applied to all existing DRL training algorithms. In this paper, we mainly evaluate the proposed GST for TD3 \cite{TD3} algorithm in Mujoco environments, which is one of the state-of-the-art DRL training algorithms. Halfcheetah-v2 and Humanoid-v2 are adopted among Mujoco environments for the GST evaluation. We used the same network configuration as the configuration utilized in \cite{TD3}. The network used in the Halfcheetah-v2 and Humanoid-v2 consists of 3 FC layers. The number of neurons in the 1st FC layer and the last FC layer is determined according to the state and action dimension of each environment. In the case of Halfcheetah-v2, the grouping and pruning of the proposed GST were only applied to the second FC layer, which accounts for 91.7\% of the total parameters. In the case of Humanoid-v2, the grouping and pruning were applied to the first and second FC layers, which account for 97.4\% of the total parameters. Network parameter initialization and training hyperparameters were the same as those introduced in \cite{TD3} for a fair comparison. All results on Mujoco environments are reported after averaging 5 random seeds' results of the Gym simulator and the network initialization.

Moreover, we verify the proposed GST on not only various RL algorithms (A2C and PPO \cite{PPO}) with various test environments (Atari Breakout and Google Research Football \cite{grf}) but also classification datasets (CIFAR-10 \cite{cifar10} and ILSVRC-2012 \cite{imagenet}) with the famous convolutional neural networks (Resnet-32 \cite{resnet} and Alexnet \cite{alexnet}). 

\subsection{Gradual Pruning vs Reward-aware Pruning}
We first verify the effectiveness of the proposed reward-aware pruning  (RWP) in order to determine the pruning methodology. To compare the performance of gradual pruning and RWP, we first apply RWP for different pruning start points (0.0M, 0.2M). After the reward and compression ratio of RWP is measured, we apply gradual pruning to have the same compression ratio as RWP and check the reward loss.

Figure~\ref{fig:resultGPvsRWP} shows the comparison results of gradual pruning and RWP. In the Halfcheetah-v2, the RWP achieved a 49.1\% average compression ratio for pruning start point 0.0M. In the Humanoid-v2, the RWP achieved a 25.0\% average compression ratio for pruning start point 0.0M. Interestingly, when gradual pruning is applied to achieve the same compression ratio as RWP, a small reward loss appears in the halfcheetah-v2, whereas a very large reward loss appears in the humanoid-v2. This is due to the training stability difference caused by distinct environments. The reward graph of halfcheetah-v2 converges early and training is stable, whereas the reward graph of  humanoid-v2 fluctuates and training is unstable. In the case of gradual pruning, target sparsity is increased even at points where training is unstable and saliency of the parameter cannot be determined, resulting in a large reward loss in humanoids. On the other hand, reward loss could be prevented by dynamically stopping pruning at the corresponding points in the case of RWP.

\subsection{GST Performance According to $B$ and $S_{shift}$}
To verify the performance of the proposed GST, we measured the reward and compression ratio on the Mujoco dataset according to block size $B$ and phase shift sparsity $S_{shift}$. The pruning start point was fixed at 0.0M, and the reward and compression ratio were measured for $B=2, 4$, and phase shift sparsity $S_{shift}=0.25, 0.5, 0.75, 1.0$.

Figure~\ref{fig:resultHCGST} shows the proposed GST measurement results on the Halfcheetah-v2 environment. For all the $B$ values, there is no training reward drop of the proposed GST when $S_{shift}$ values of 0.5 or less were used, and a negligible reward drop occurred even at higher $S_{shift}$ values. The maximum compression ratio of 74.9 \% was obtained under the condition of $S_{shift}=1.0$ and $B=4$, which is 25 \%p higher than when only RWP was applied. 

Figure~\ref{fig:resultHUGST} shows the proposed GST measurement results on the Humanoid-v2 environment. In the case of humanoid, there is no result for $S_{shift}$ = 1.0 because sparsity of 75\% or more was not achieved. When $B=2$ was used, a similar level of baseline reward could be achieved for all $S_{shift}$ values. When $B=4$ was used, a large reward drop occurred for $S_{shift}=0.5, 0.75$. The maximum compression ratio of 61.9 \% was obtained without reward drop under the condition of $S_{shift}=1.0$ and $B=2$, which is 36.9 \%p higher than when only RWP was applied.

In both environments, we can observe two tendencies: 1) Even in the case of GST with a large $B$, a similar level of reward compared with baseline achieved for the early iterations. 2) After the phase transition occurs, the reward of GST gradually recovers to the baseline reward level. Therefore, we can obtain both a high compression ratio and a high reward by releasing block-circulant at an appropriate time. 

\begin{figure}
\begin{center}
    \begin{subfigure}{0.4\textwidth}
        \includegraphics[width=\linewidth]{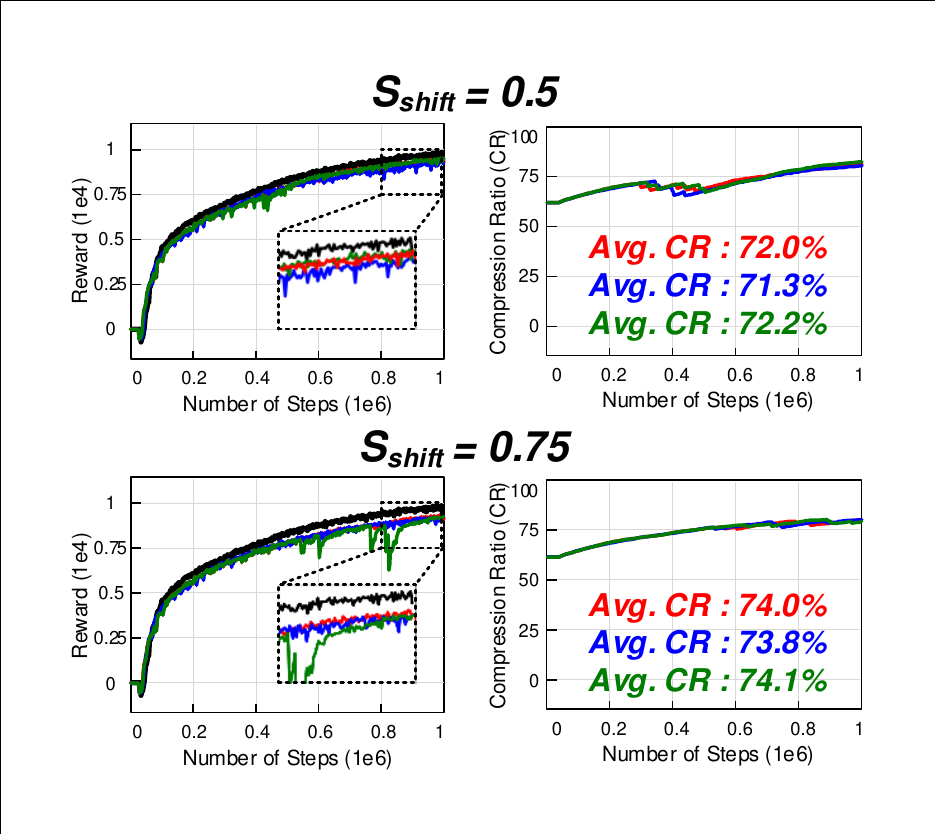}
        \caption{Result on mujoco halfcheetah-v2 with TD3 \cite{TD3}} \label{fig:8a}
    \end{subfigure}
    \begin{subfigure}{0.4\textwidth}
        \includegraphics[width=\linewidth]{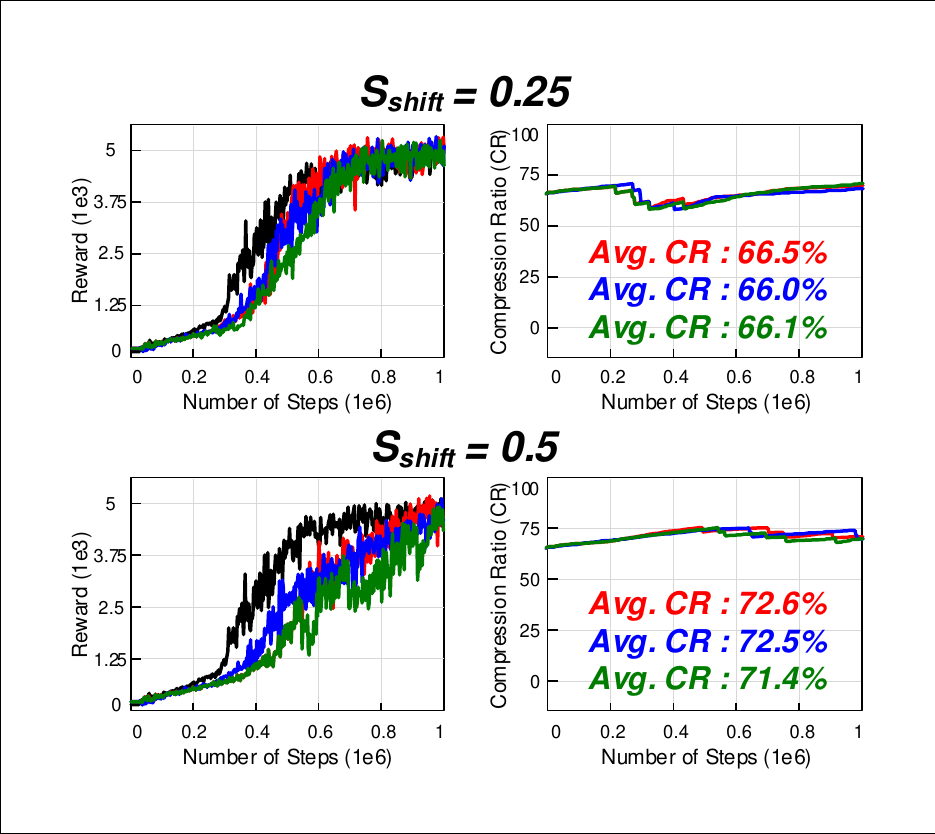}
        \caption{Result on mujoco humanoid-v2 with TD3 \cite{TD3}} \label{fig:8b}
    \end{subfigure}
\end{center}
\vspace*{-3mm}
   \caption{Reward and compression ratio measurement results according to the three block size conversion methods. Black lines are baseline result, red lines are results of block4 friendly block2 method, blue lines are results of block2 friendly block4 method, green lines are results of projection method.}
\label{fig:resultB4B2}
\end{figure}

\subsection{Performance of Block Size Conversion Methods}
Figure~\ref{fig:resultB4B2} shows the GST measurement result with 3 different block size conversion methods on the halfcheetah-v2 and humanoid-v2. For all $S_{shift}$ values in both environments, the projection method showed the worst training reward, and the block4 friendly block2 method showed the highest training reward. In the projection method, a large fluctuation in reward occurred due to a sudden change in parameter value for each phase transition, which made training unstable. On the other hand, the methods using the friendly matrix enabled stable training, and the same level of training reward as the block circulant method was obtained. All three methods have similar average compression ratios. In the case of Halfcheetah-v2, there is no advantage of block conversion methods because it is possible to utilize block4 in the entire training. In the case of Humanoid-v2, the maximum compression ratio of 66.5 \% was obtained without reward drop under the condition of $S_{shift}=0.25$ and block4 friendly block2 method, which is 4.6 \%p higher than result of the GST without block conversion method.

\begin{figure}
\begin{center}
    \begin{subfigure}{0.4\textwidth}
        \includegraphics[width=\linewidth]{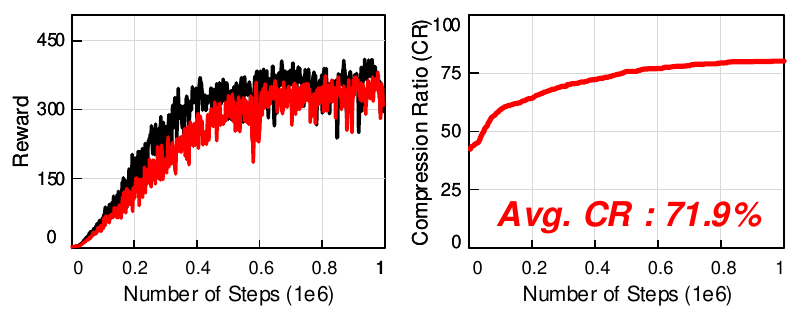}
        \caption{Result on atari breakout with A2C} \label{fig:9a}
    \end{subfigure}
    \begin{subfigure}{0.4\textwidth}
        \includegraphics[width=\linewidth]{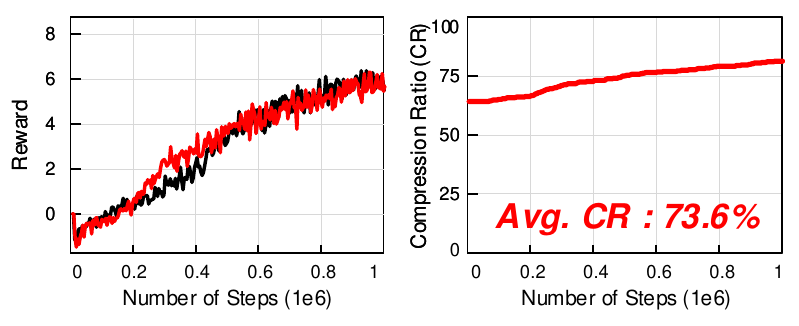}
        \caption{Result on google research football with PPO} \label{fig:9b}
    \end{subfigure}
\end{center}
\vspace*{-3mm}
   \caption{Reward and compression ratio measurement results of GST on various DRL benchmark. Black lines are baseline result, red lines are GST result.}
\label{fig:resultAtraiGRFGST}
\end{figure}
\begin{figure}
\begin{center}
    \begin{subfigure}{0.4\textwidth}
        \includegraphics[width=\linewidth]{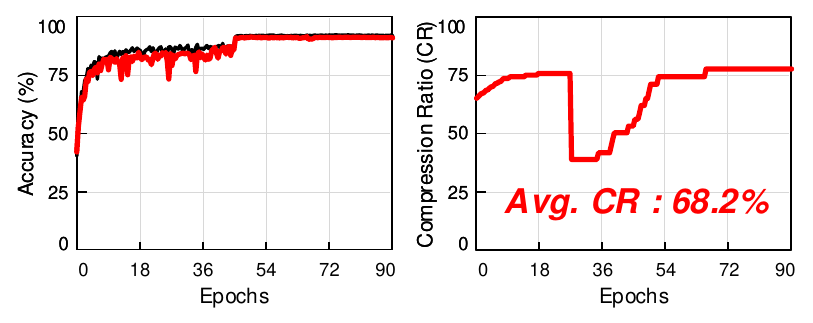}
        \caption{Cifar-10 training result with ResNet-32} \label{fig:10a}
    \end{subfigure}
    \begin{subfigure}{0.4\textwidth}
        \includegraphics[width=\linewidth]{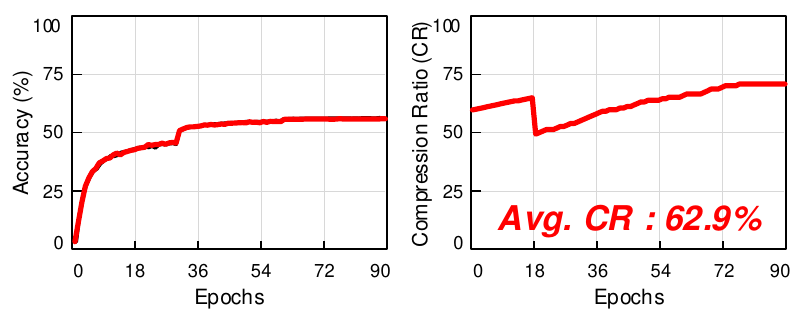}
        \caption{ImageNet training result with Alexnet} \label{fig:10b}
    \end{subfigure}
\end{center}
\vspace*{-3mm}
   \caption{Accuracy and compression ratio measurement results of GST on classification network training. Black lines are baseline results, red lines are GST results.}
\label{fig:resultCifar10ImageNetGST}
\end{figure}

\subsection{Verification GST on Various DRL Benchmark}
We tested the proposed GST on a more diverse DRL benchmark to verify that it can operate regardless of the layer types (convolution layer, FC layer), RL algorithms, and task types. Figure~\ref{fig:resultAtraiGRFGST} shows the training results of Atari Breakout with A2C and training results of Google Research Football with PPO. In the Atari Breakout training, a network configuration and training hyperparameter were the same as those introduced in \cite{pytorchrl}, and we run 5 random seed and average the results. The network consists of 3 convolutional layers and 2 FC layers, and GST was applied for 99\% of the weight excluding the first and last layers. In this case, the highest average compression ratio of 71.9\% was achieved without a reward drop when the $B=2$ and $S_{shift}=1.0$. In the Google Research Football training, we utilized the same PPO network configuration and training parameters as \cite{grf}, and we run 3 random seed and average the results. We applied GST to 94.3\% of parameters excluding the first and last layers. We achieved an average compression ratio of 73.6 \% without reward loss when the $B=4$ and $S_{shift}=1.0$. 

\subsection{Verification GST on Classification Networks}
To verify the generality of the proposed GST, we applied the GST to the famous classification benchmark. Figure~\ref{fig:resultCifar10ImageNetGST} shows the results of training cifar-10 with ResNet-32 and training ilsvrc-2012 with Alexnet. We utilized the same network configuration and the same hyperparameters as the pytorch official implementation for both experiments. In the cifar-10 training, GST was applied to 94.8\% of parameters including layer 12 to layer 31 of ResNet-32. The proposed GST achieved an average compression ratio of 68.2 \% when the $B=4$ and $S_{shift}=0.5$. The accuracy was 91.4 \%, which was 0.8 \% lower than baseline accuracy of 92.2 \%. In the case of ilsvrc-2012 training, GST was applied to 87.4\% of parameters including FC5 and FC6 of Alexnet. The proposed GST achieved an average compression ratio of 62.9 \% when the block4 friendly block2 method and $S_{shift}=0.25$ utilized. The accuracy was 55.8 \%, which was 0.4 \% lower than baseline accuracy of 56.2 \%.

\section{Conclusion and Future Work}
In this paper, we propose a novel weight compression method for DRL training acceleration, group sparse training (GST). To overcome the low compression ratio at the early iteration and unstable training due to the fixed sparsity schedule, GST selectively utilizes block-circulant compression for high compression ratio and dynamically adapt target sparsity through reward-aware pruning for stable training. Thanks to the features, GST achieves a 25 \%p $\sim$ 41.5 \%p higher average compression ratio than the previous iterative pruning method without reward degradation. To the best of our knowledge, this paper is the first research to compress a DRL network training procedure through pruning without a pre-trained teacher network. In the future, we plan to work on an automatic hyperparameters search algorithm for finding optimized pruning threshold and phase shift sparsity online.

{\small
\bibliographystyle{unsrt}
\bibliography{gst_arxiv}
}

\end{document}